# Capsule Neural Networks as Noise Stabilizer for Time Series Data

Soyeon Kim[O1]  Jihyeon Seong[1]  Hyunkyung Han[1]  Jaesik Choi[12]

[1] Korea Advanced Institute of Science and Technology  [2] INEEJI

### abstract

Capsule Neural Networks (CapsNets) utilize capsules, which bind neurons into a single vector and learn position-equivariant features, which makes them more robust than original Convolutional Neural Networks (CNNs). CapsNets employ an affine transformation matrix and dynamic routing with coupling coefficients to learn robustly. In this paper, we investigate the effectiveness of CapsNets in analyzing highly sensitive and noisy time series sensor data. To demonstrate CapsNets' robustness, we compare their performance with original CNNs on electrocardiogram (ECG) data, a medical time series sensor data with complex patterns and noise. Our study provides empirical evidence that CapsNets function as noise stabilizers, as investigated by manual and adversarial attack experiments using the fast gradient sign method (FGSM) and three manual attacks, including offset shifting, gradual drift, and temporal lagging. In summary, CapsNets outperform CNNs in both manual and adversarial attacked data. Our findings suggest that CapsNets can be effectively applied to various sensor systems to improve their resilience to noise attacks. These results have significant implications for designing and implementing robust machine-learning models in real-world applications. Additionally, this study contributes to the effectiveness of CapsNet models in handling noisy data and highlights their potential for addressing the challenges of noise data in time series analysis.

## 1. Introduction

Capsule Neural Network (CapsNet) [5, 6] architecture is highly effective for maintaining object invariance in position-equivariant data, such as images captured from different viewpoints [1]. Observed data by sensors, however, includes unnoticeable noise despite devoting efforts to calibration and maintenance of sensors due to unexpected environmental changes, low battery state, and others [2]. The faulty data must be stabilized to enhance model prediction performance. CapsNet achieves view-equivariance by utilizing an affine transformation matrix and dynamic routing, which can also effectively compensate for noisy data.

In this study, we hypothesize that CapsNets are resistant to noisy time-series sensor data and attacks caused by defective sensors in the real world. To test our hypothesis, we conducted experiments using electrocardiogram (ECG) data and compared the performance of CapsNets with that of original CNNs. Our results confirm that even with added noise from attacks, CapsNets outperformed the original CNNs, providing evidence to support the hypothesis [1, 5, 6].

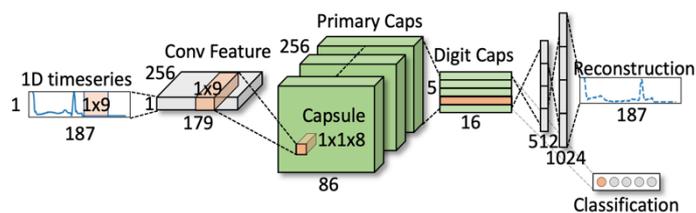

Figure 1  Dynamic routing capsule neural network on One Dimensional TimeSeries Data

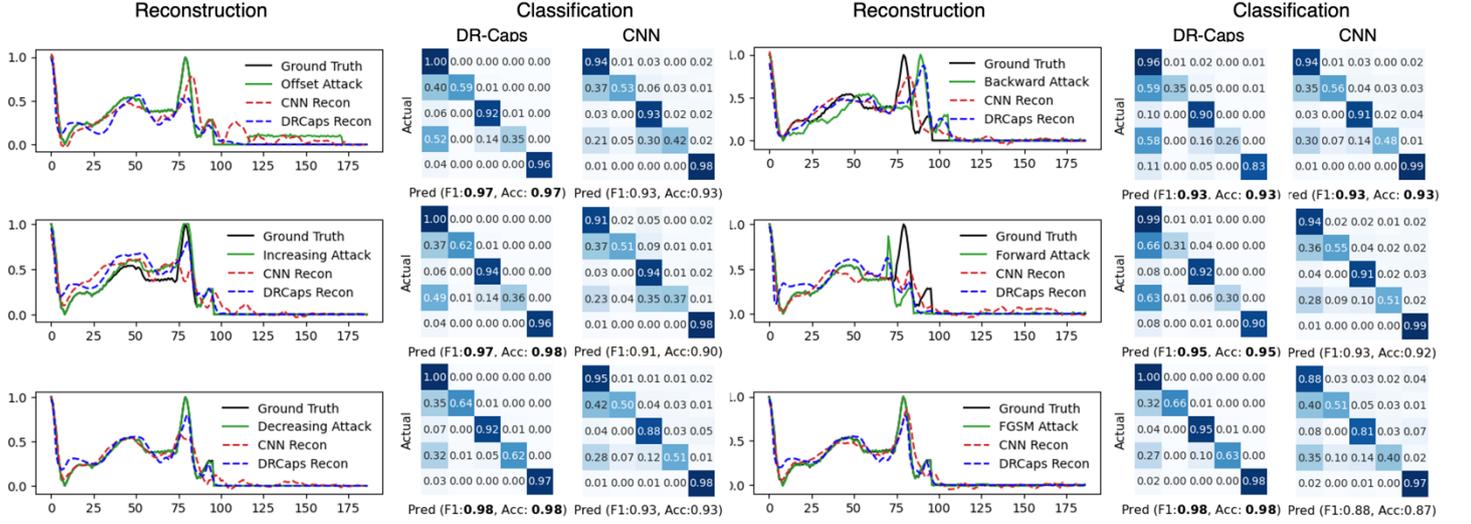

Figure 2  Reconstruction and classification performance of DR-Caps and CNN on attacked data

## 2. Background

Dynamic Routing CapsNets (DR-CapsNets) consist of two distinct components with original CNNs; the affine transformation matrix and dynamic routing. The affine transform matrix is resistant to affine-type noise attacks, while dynamic routing incorporates a bi-directional attention mechanism that determines the attention levels of lower-layer temporal features relative to the target capsules of the upper layer [4, 7]. As a result, DR-CapsNets enhance fundamental temporal features while inhibiting noisy features that appear less frequently in the training data. We assume the features of fundamental signals are invariant even with added noise. The invariant knowledge about signals is contained in the affine transformation matrices that predict the pose of a whole signal from the pose of a part (capsule) since the capsule includes a part of temporal features with the length of a receptive field (Fig. 3).

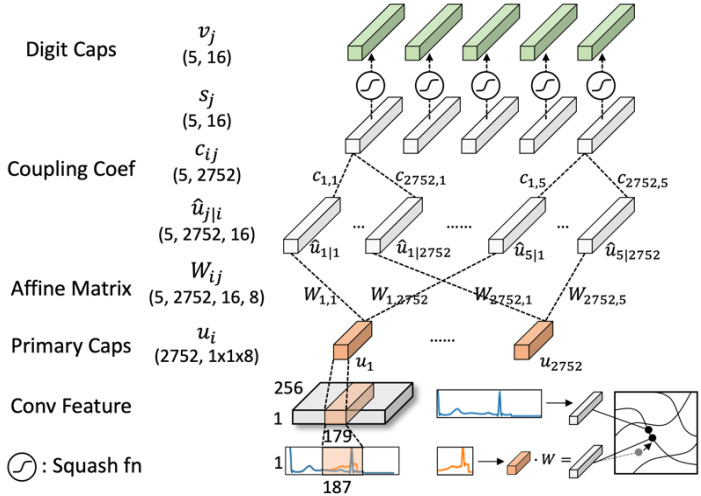

Figure 3 The role of affine matrix as sensor stabilizer

## 3. Method

In this study, we conducted experiments to evaluate the robustness of the deep learning model against various types of noise attacks, including manual and adversarial attacks. To simulate manual noise attacks, we defined three types of noise as referred from the research on types of sensor defects in the real world [2]: offset, increasing/decreasing drift, and forward/backward lagging. To add randomness to each manual noise attack, we utilized a stochastic differential equation borrowed from the stock price process, known as the Black-Scholes model (Eq. 1) [3, 4]. We called this process the *noise move simulation* and estimated the noise distribution the next time using a log-normal distribution.

$$\frac{dS}{S} = \mu dt + \sigma db \quad \text{(Eq. 1)}$$

For adversarial attacks, we utilized the Fast Gradient Sign Method (FGSM) [8] which modifies the input by taking a small step toward the gradient sign of the loss function with respect to the input (Eq. 2). The use of manual and adversarial noise attacks allowed us to thoroughly evaluate the robustness of the deep learning model against a broad range of potential noisy scenarios that could occur in real-world applications as shown in Figure 3.

$$x' = x + \alpha \, \text{sign}(\nabla_x L(\theta, x, y)) \quad \text{(Eq. 2)}$$

By simulating both manual and adversarial noise attacks, we demonstrated our model's effectiveness in handling various types of noise and highlighted its potential for improving the accuracy and robustness of time series sensor data analysis.

| w/o Attack | Accuracy | F1-Score |
|---|---|---|
| CNN | 94.17% | 94.81% |
| DR-Caps | 98.22% | 98.13% |

Table 4 Classification performance without attack

## 4. Experiments

To evaluate the effectiveness of our proposed method, we conduct a series of experiments on an ECG dataset of real-world time series data measured by electrical sensors. We compare the performance of DR-CapsNets against CNN under several noise baselines, including affine transform noise (lagging, gradually increasing and decreasing, offsets) and adversarial attacked noise. We compare DR-CapsNet to the original CNN sharing the same spec.

- Dataset: ECG has smooth-shaped medical heartbeat signals with five types of heartbeat classification tasks.
- Experimental Setup: We perform three types of noise attacks: offset, gradual drift, and lagging, as described in the Method section. We generate 100 manually crafted noisy samples for each type of attack, with varying noise levels. We also perform FGSM attack with the value of $\alpha = 0.01$.
- Evaluation Metrics: We evaluate the performance of our method using several metrics. Our primary metric is accuracy, which measures the percentage of correctly classified samples. We also calculate the f1-score, which addresses the class imbalance problem in a dataset.

As shown in Figure 2, CapsNet outperforms with four different noise attacks.

## 5. Results

The results of our experiments demonstrate the effectiveness of DR-CapsNets in handling noise attacks on ECG data. Our method outperformed CNN in all three noise baselines, including affine transform noise and adversarial attacked noise. Specifically, DR-CapsNets achieved an accuracy of 98.7%, 97.5%, 97.2%, and 94.8% for the offset, gradually increasing/decreasing, and forward/backward lagging attacks, respectively. Moreover, our method also outperformed CNN in terms of the f1-score, which indicates its robustness against class imbalance in the dataset.

## 6. Conclusions

Our study demonstrates that CapsNets can effectively analyze highly sensitive and noisy time series sensor data such as ECG data. Our findings show that CapsNets function as noise stabilizers and are more robust than original CNNs in handling manual and adversarial attack data. Specifically, the DR-CapsNets method outperformed CNN in both accuracy and f1-score in all three noise baselines, including affine transform noise and adversarial attacked noise. These results highlight the potential of CapsNets, for improving the accuracy and robustness of ECG heartbeat classification tasks in real-world scenarios. Our study also contributes to the effectiveness of CapsNet models in handling noisy data and highlights their potential for addressing the challenges of noisy data in time series analysis. These findings have significant implications for designing and implementing robust machine-learning models in various sensor systems and real-world applications.